\documentclass[11pt]{article}
\usepackage{float}
\usepackage{amsmath}
\usepackage{amssymb}

\usepackage[preprint]{acl}

\usepackage{times}
\usepackage{latexsym}

\usepackage[T1]{fontenc}

\usepackage[utf8]{inputenc}

\usepackage{microtype}

\usepackage{inconsolata}

\usepackage{graphicx}
\usepackage{float}
\usepackage{dblfloatfix}
\usepackage{booktabs}
\usepackage{tabularx}
\usepackage{colortbl}   
\usepackage{array}
\usepackage{multirow}
\usepackage{xcolor}

\newcommand{\cardelta}[1]{\textcolor{teal}{\scriptsize{(#1)}}}
\newcommand{\carneg}[1]{\textcolor{red}{\scriptsize{(#1)}}}
\newcommand{\baseDelta}[1]{\textcolor{black}{\scriptsize{(#1)}}}
\usepackage{algorithm}
\usepackage{algpseudocode}
\newcommand{\masktoken}{\mbox{\textsc{[mask]}}}
\usepackage{tikz}
\usetikzlibrary{arrows.meta,positioning,calc,fit,decorations.pathreplacing}

\title{\textsc{CARVE}: Verified Expansion for Variable-Length Generation in\\ Diffusion Language Models}

\author{
\textbf{Wail Bouhedja\textsuperscript{1,2}},
 \textbf{Amr Mohamed\textsuperscript{1,3}},
 \textbf{Guokan Shang\textsuperscript{1}}
\\
\\
 \textsuperscript{1}MBZUAI,
 \textsuperscript{2}Sorbonne Université,
 \textsuperscript{3}Ecole Polytechnique
\\
}

\begin{document}
\newcommand{\methodname}{CARVE}
\maketitle
\begin{abstract}
Masked diffusion language models predict tokens from a partially observed response canvas, enabling bidirectional conditioning and parallel token refinement. Yet standard masked-diffusion decoders use a rigid inference interface: the number of masked positions allocated to the answer is fixed before generation begins. Choosing this length is difficult. A short canvas can truncate reasoning or code, while a long canvas wastes computation and can perturb denoising. We introduce \methodname{} (\textbf{C}ounterfactual-\textbf{A}ware \textbf{R}eveal with \textbf{V}erified \textbf{E}xpansion), a training-free variable-length algorithm for masked diffusion LMs. Starting from a shorter canvas, \methodname{} can grow the response during decoding by inserting additional \texttt{[MASK]} positions. Rather than keeping every insertion, \methodname{} tests a candidate expanded canvas and asks a counterfactual question: would the model make similar predictions for the unresolved positions in the original canvas if the extra masked space were present? The inserted masks are kept only when they induce low Jensen--Shannon (JS) divergence on aligned unresolved positions. This makes length growth a verified stability decision rather than a pure confidence heuristic. \methodname{} applies without retraining to both full-canvas and blockwise diffusion decoders. Across code generation and mathematical reasoning benchmarks, \methodname{} consistently improves average performance over fixed-length baselines across all evaluated model families. Crucially, \methodname{} achieves these accuracy gains while reducing inference cost, reaching half the FLOPs of fixed-length decoding in some settings. 
\end{abstract}

\section{Introduction}

Autoregressive language models \cite{llm} have a simple interface for open-ended generation: they emit one token at a time and stop when an end-of-sequence token is produced. Masked diffusion language models \citep{austin2021structured, campbell2022continuous, zheng2023reparam, lou2023discrete, sahoo2024simple, shi2024simplified}, by contrast, generate by iteratively denoising a canvas of masked tokens. This paradigm enables attractive properties such as parallel token updates, bidirectional conditioning, and arbitrary-order refinement, and has recently scaled to competitive instruction-following models such as LLaDA \cite{llada} and Dream \cite{dream}. However, it exposes a basic inference problem: the generation length must typically be fixed before decoding begins.

This fixed-canvas assumption is poorly matched to realistic generation. If the canvas is too short, the model may truncate code, reasoning, or explanations. If the canvas is too long, inference wastes computation on masked positions that may not be needed for the final answer, and can even harm quality by forcing the model to denoise beyond the useful response \cite{length_aware_dlm}. The appropriate length is also instance-dependent: two prompts from the same task may require very different response lengths, even when they use the same decoding setup. As a result, variable-length decoding is not a minor engineering detail, but a central obstacle to making diffusion LMs practical for open-ended generation.

A natural solution is to adapt the canvas during denoising. Prior work such as DAEDAL \cite{daedal} proposes training-free length expansion for diffusion language models by using internal confidence signals to decide when to allocate additional masked tokens. This shows that response length need not be fixed before decoding
begins and can instead be adjusted during inference. Yet expansion introduces a second question that is easy to overlook: adding masks changes the denoising problem itself. In a bidirectional masked diffusion model, inserting new masked positions may perturb the model's predictions at other still-unresolved positions. An appropriate length expansion rule should therefore ask both whether additional space might be useful and whether inserting that space leaves the existing predictions sufficiently stable.

We introduce \methodname{} (\textbf{C}ounterfactual-\textbf{A}ware \textbf{R}eveal with \textbf{V}erified \textbf{E}xpansion), a training-free variable-length decoding algorithm for masked diffusion language models. \methodname{} augments a standard diffusion sampler with a verified expansion step. Before certain reveal steps, CARVE proposes inserting a number of additional masked tokens into the current canvas. It then runs the model once on the expanded canvas and compares the predictive distributions of the original and expanded canvases at aligned still-masked positions. The insertion is accepted only when the mean Jensen--Shannon divergence between these distributions remains below a fixed threshold. Otherwise, the expanded branch is rejected and we continue with the original canvas. In this way, the canvas grows only when the proposed expansion leaves the predictive distributions at aligned unresolved positions sufficiently stable under this criterion.

This turns length expansion into a stability check rather than a pure confidence heuristic. Each proposed insertion asks a local counterfactual question: would the model make essentially the same predictions for the unresolved tokens if the canvas contained additional space? When the answer is yes, the added masks can be committed without substantially changing the denoising state. When the answer is no, the proposal is treated as a destabilizing edit and discarded. This criterion differs from expansion rules based only on absolute confidence or end-of-sequence (EOS) token behavior \cite{rho_eos}: it measures the effect of the insertion on the model's remaining predictive state. The same principle applies across different masked diffusion backends; in this work, we instantiate \methodname{} for both full-canvas Dream decoding and blockwise LLaDA decoding.

Our experiments evaluate \methodname{} on code and mathematical reasoning benchmarks across different dLLMs. We compare against fixed-length baseline decoding and DAEDAL \cite{daedal}, the closest training-free variable-length baseline. Across models, \methodname{} improves average performance over fixed-length baseline decoding. Our analysis of model-forward FLOPs further shows that verified expansion does not simply trade additional computation for accuracy: adaptive reveal and EOS cropping often offset the cost of branching. Consequently, \methodname{} provides a dual advantage, improving average task accuracy while yielding up to roughly half as many FLOPs as fixed-length decoding in some settings. Code is publicly available\footnote{\url{https://github.com/wailji/CARVE}}. 
Our contributions are:
\begin{itemize}
    \item We propose \methodname{}, a training-free variable-length algorithm that verifies each proposed canvas expansion by measuring Jensen--Shannon divergence on aligned unresolved positions.
    \item We show that verified expansion provides a model-agnostic mechanism for masked diffusion LMs, applying the same algorithm to both full-canvas Dream decoding and blockwise LLaDA decoding.
    \item We demonstrate that \methodname{} improves average performance over fixed-length baselines across code and mathematical reasoning benchmarks on three diffusion LMs, while often reducing inference FLOPs relative to fixed-length decoding.
\end{itemize}

\section{Related Work}
\label{sec:related_work}

\paragraph{Discrete diffusion language models.}
Masked diffusion large language models (dLLMs) generate sequences by iteratively reversing categorical corruption processes, typically relying on absorbing-state masking to denoise positions in parallel \citep{austin2021structured, campbell2022continuous, zheng2023reparam, lou2023discrete, sahoo2024simple, shi2024simplified,mohamed-etal-2026-fast}. Recent large-scale models, such as LLaDA and Dream, demonstrate that this paradigm scales effectively to complex instruction-following and reasoning tasks \citep{llada, dream}. However, because these models predict and reveal subsets of tokens across a predefined masked canvas, they inherently impose a strict length constraint before decoding begins.

\paragraph{Training-time variable-length generation in dLLMs.}
To overcome the fixed-length bottleneck, several approaches modify the underlying generative formulation or state space. Existing methods implement dynamic expansion and contraction for code infilling \citep{dreamon}, formulate generation via explicit insertion and deletion edits \citep{havasi2025editflows}, or jointly denoise token identities and positional configurations \citep{zhang2025flexible, kim2025any}. While these strategies successfully enable dynamic length adjustment, they require specialized architectural modifications or costly retraining. \methodname{}, by contrast, applies directly to pretrained masked dLLMs.

\paragraph{Training-free variable-length decoding in dLLMs.}
Inference-time methods bypass retraining by dynamically adjusting the canvas length during decoding. Prior approaches trigger length changes using scalar internal confidence cues \citep{daedal}, length-regularized candidate scoring \citep{lrdllm}, or implicit end-of-sequence (EOS) token densities \citep{rho_eos}. \methodname{} instead frames length expansion as a counterfactual stability test. It computes the mean Jensen--Shannon divergence between the base and expanded predictive distributions at aligned unresolved positions and accepts the expansion when this mean falls below $\tau_{\mathrm{JS}}$. This criterion measures local predictive stability; it does not guarantee positionwise preservation, semantic correctness, or output safety.

\section{Methods}
\label{sec:methods}

\begin{figure*}[t]
    \centering
    \includegraphics[width=\textwidth]{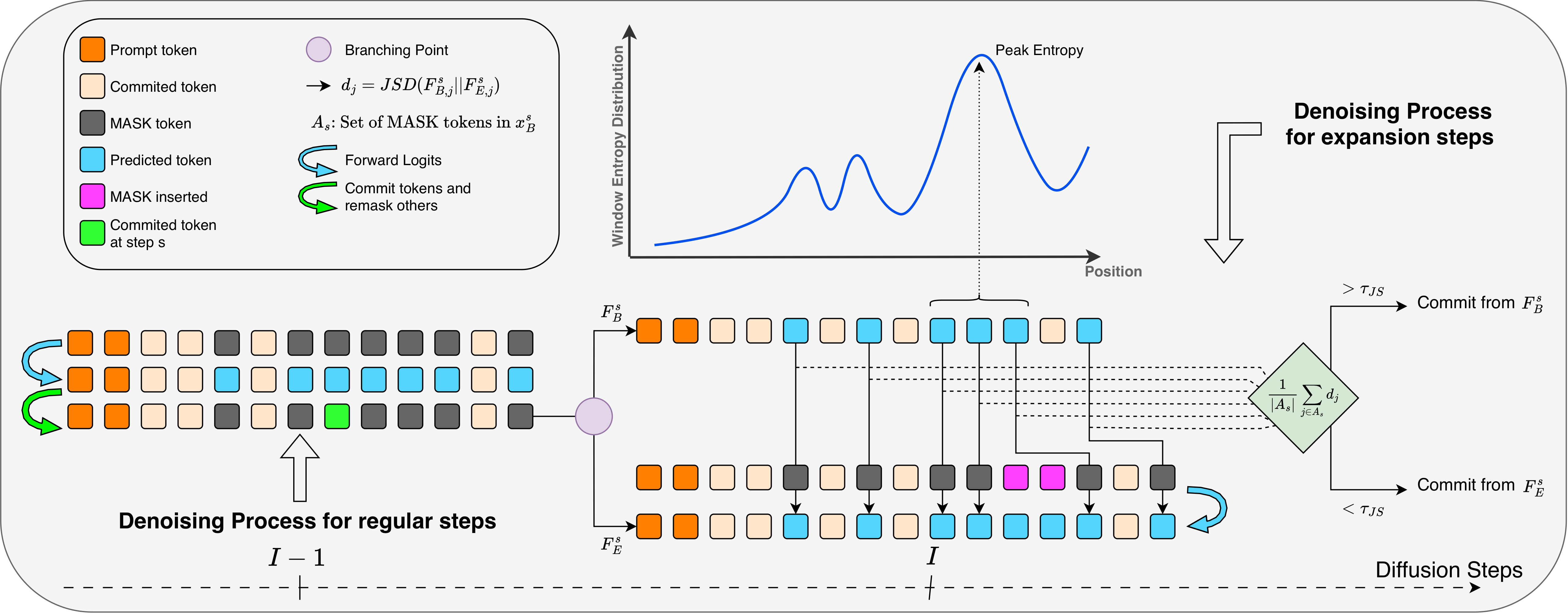}
    \caption{Overview of \methodname{}. Regular steps predict all masked positions and commit a subset of tokens. Expansion steps insert additional \masktoken{} tokens near a high-uncertainty region and compare the base and expanded canvases using JS divergence. The expanded canvas is kept only when the mean JS divergence on aligned unresolved positions falls below the acceptance threshold.}
    \label{fig:carve_overview}
\end{figure*}

In this section, we introduce \methodname{}, a training-free variable-length decoding algorithm for masked diffusion language models. Figure~\ref{fig:carve_overview} provides an overview, and Algorithm~\ref{alg:carve} summarizes the complete decoding procedure.

\subsection{Preliminaries: Masked Discrete Diffusion}
\label{sec:preliminaries}

Masked diffusion language models generate text by denoising discrete token sequences. Given a clean response $x_0=(x_{0,1},\ldots,x_{0,L})\in\mathcal{V}^{L}$, the forward process gradually replaces tokens with a special mask token $\masktoken$. Since masked tokens remain masked at all later timesteps, this is an \emph{absorbing} process. At inference time, decoding starts from a masked canvas and progressively commits predicted tokens.

\paragraph{Forward absorbing process.}
The forward process is a Markov chain
\begin{equation}
    q(x_{1:T}\mid x_0)
    =
    \prod_{t=1}^{T} q(x_t\mid x_{t-1}),
\end{equation}
with transitions that factorize over token positions. For each position $i$, $\masktoken$ is absorbing:
\begin{equation}
    q(x_{t,i}=\masktoken \mid x_{t-1,i}=\masktoken)=1.
\end{equation}
If $x_{t-1,i}\neq\masktoken$, then
\begin{equation}
x_{t,i}
=
\begin{cases}
x_{t-1,i}, & \text{with probability } 1-\beta_t,\\
\masktoken, & \text{with probability } \beta_t.
\end{cases}
\label{eq:absorbing_transition}
\end{equation}
We assume that the corruption schedule satisfies $\bar{\alpha}_T=0$, so the terminal state is fully masked: $x_T=\masktoken^L$. Let $\bar{\alpha}_t=\prod_{r=1}^{t}(1-\beta_r)$ denote the token survival probability. The marginal corruption process is
\begin{equation}
x_{t,i}
=
\begin{cases}
x_{0,i}, & \text{with probability } \bar{\alpha}_t,\\
\masktoken, & \text{with probability } 1-\bar{\alpha}_t.
\end{cases}
\label{eq:forward_marginal}
\end{equation}

\paragraph{Denoising model.}
Given a prompt $x_{\mathrm{p}}$ and a partially masked response $x_t$, the model predicts logits over the vocabulary at every response position:
\begin{equation}
    F_t=f_\theta(x_{\mathrm{p}},x_t,t).
\end{equation}
The token distribution at position $i$ is
\begin{equation}
    p_{t,i}=\mathrm{softmax}(F_{t,i}),
\end{equation}
where $F_{t,i}$ denotes the logits at position $i$.
We write the resulting clean-token predictor in factorized form:
\begin{equation}
    p_\theta(x_0\mid x_{\mathrm{p}},x_t,t)
    =
    \prod_{i=1}^{L}
    p_\theta(x_{0,i}\mid x_{\mathrm{p}},x_t,t).
\end{equation}
Although the output distribution factorizes over response positions, each factor is computed from the full partially masked canvas, allowing bidirectional conditioning on all visible tokens.

\paragraph{Training objective.}
The model is trained to reconstruct the original tokens at masked positions. Let $M_t=\{i:x_{t,i}=\masktoken\}$. The masked denoising loss is

\begin{equation}
\!
\begin{aligned}
\mathcal{L}(\theta)
&=
\mathbb{E}_{(x_{\mathrm{p}},x_0)\sim\mathcal{D}}
\mathbb{E}_{t\sim\mathcal{U}\{1,\ldots,T\}}
\mathbb{E}_{x_t\sim q(\cdot\mid x_0,t)}\\&
\left[
-\sum_{i\in M_t}
\log p_\theta
\left(
x_{0,i}\mid x_{\mathrm{p}},x_t,t
\right)
\right].
\end{aligned}
\label{eq:masked_denoising_loss}
\end{equation}

\paragraph{Decoding with partial reveal.}
At inference time, the response length $L$ is chosen in advance and the decoder initializes a fully masked response canvas,
\begin{equation}
    x^{(0)}=\masktoken^{L}.
\end{equation}
Let $\{\tau_s\}_{s=0}^{T}$ denote the reverse denoising schedule, with $\tau_0=T$ corresponding to the maximally corrupted state and $\tau_T=0$ to the clean state. At decoding step $s$, the model predicts token distributions for all response positions at diffusion time $\tau_s$. A reveal rule then selects a subset of the currently masked positions to commit. Let
\begin{equation}
m_s=
\left|
\left\{i:x_{\mathrm{C},i}^s=\masktoken\right\}
\right|
\end{equation}
denote the number of unresolved positions in the chosen branch. The canvas is updated as
\begin{equation}
x_i^{(s+1)}
=
\begin{cases}
\hat{x}_i^{(s)}, & i\in R^{(s)},\\
x_i^{(s)}, & i\notin R^{(s)},
\end{cases}
\end{equation}
where $\hat{x}_i^{(s)}$ is obtained from the model distribution, either greedily or by sampling. Different diffusion models define different reveal rules. In this work, we keep the trained denoising model fixed and modify only the decoding procedure.

\subsection{\methodname{}: Verified Canvas Expansion}
\label{sec:carve}

\methodname{} is a training-free decoding method for masked diffusion language models. It keeps the pretrained denoising model and each backend's position-ranking criterion fixed while modifying inference through verified canvas expansion, an adaptive reveal-count schedule, and EOS-triggered cropping. \methodname{} branches into a candidate expanded canvas, measures the mean distribution shift at aligned unresolved positions, and reveals tokens from the branch selected by the verification criterion.

Let $x^s$ be the response canvas at decoding step $s$, and let $L_s$ be its current length. Standard masked diffusion decoding fixes this length before generation. In contrast, \methodname{} starts from an initial length $L_0$ and allows the canvas to grow up to a maximum length $L_{\max}$. At each step, we first define the current canvas as the \emph{base branch}:
\begin{equation}
    x_{\mathrm{B}}^s = x^s.
\end{equation}
The model is evaluated on this branch to obtain base logits:
\begin{equation}
    F_{\mathrm{B}}^s
    =
    f_\theta(x_{\mathrm{p}},x_{\mathrm{B}}^s,\tau_s).
\end{equation}

\paragraph{Uncertainty-guided branching.}
Before each potential canvas expansion, \methodname{} selects where an inserted span of masks may be most useful. It inserts new masks near the most uncertain region of the current canvas, using high predictive uncertainty as a heuristic for where additional capacity may be beneficial.
\begin{equation}
    p_{\mathrm{B},j}^s
    =
    \mathrm{softmax}(F_{\mathrm{B},j}^s).
\end{equation}
The uncertainty of position $j$ is measured by entropy:
\begin{equation}
    H_j^s
    =
    -\sum_{v\in\mathcal{V}}
    p_{\mathrm{B},j}^s(v)
    \log p_{\mathrm{B},j}^s(v).
\end{equation}
Let $W$ be an even window size and let $h=W/2$. We choose an anchor position $c_s$ whose local window has the largest total entropy:
\begin{equation}
    c_s
    =
    \arg\max_{c\in\mathcal{C}_s}
    \sum_{j=c-h+1}^{c+h} H_j^s,
\end{equation}
where $\mathcal{C}_s$ is the set of valid anchor positions whose window lies inside the current response canvas. Intuitively, $c_s$ marks the region where the model is least certain about how to complete the response.

Let
\begin{equation}
    k_s=\min(k,L_{\max}-L_s)
\end{equation}
be the number of mask tokens that can still be inserted. We then branch out by inserting these $k_s$ masks immediately to the right of the anchor position $c_s$:
\begin{equation}
    x_{\mathrm{E}}^s
    =
    \operatorname{Insert}(x_{\mathrm{B}}^s,c_s,k_s),
\end{equation}
where $\operatorname{Insert}$ preserves all existing tokens and inserts $k_s$ new mask tokens immediately after position $c_s$. This gives an \emph{expanded branch}. The model is evaluated once on this expanded canvas:
\begin{equation}
    F_{\mathrm{E}}^s
    =
    f_\theta(x_{\mathrm{p}},x_{\mathrm{E}}^s,\tau_s).
\end{equation}

\paragraph{Aligning base and expanded branches.}
The inserted masks shift every original position to the right of the anchor. To compare the base and expanded branches, we align each original response position with its corresponding position in the expanded branch. For compactness, response indices omit the prompt offset. The aligned index of an original position $j$ is
\begin{equation}
    a_s(j)
    =
    j+k_s\mathbf{1}[j>c_s].
\end{equation}
Thus, positions up to and including $c_s$ keep their index, while positions after $c_s$ shift by $k_s$ slots. The newly inserted masks are excluded from the comparison because they have no counterpart in the base branch.

We verify the expansion only on unresolved positions that already existed in the base canvas:
\begin{equation}
    \mathcal{A}_s
    =
    \{j:x_{\mathrm{B},j}^s=\masktoken\}.
\end{equation}
Committed tokens are excluded because their values are already fixed.

\paragraph{JS verification score.}
For every aligned unresolved position $j\in\mathcal{A}_s$, we compare the model's predictive distribution before and after insertion:
\begin{equation}
    p_j
    =
    \mathrm{softmax}(F_{\mathrm{B},j}^s),
\end{equation}
\begin{equation}
    q_j
    =
    \mathrm{softmax}(F_{\mathrm{E},a_s(j)}^s).
\end{equation}
Let
\begin{equation}
    m_j=\frac{1}{2}(p_j+q_j).
\end{equation}
The local Jensen--Shannon divergence is
\begin{equation}
    d_j
    =
    \frac{1}{2}\mathrm{KL}(p_j\|m_j)
    +
    \frac{1}{2}\mathrm{KL}(q_j\|m_j).
\end{equation}
The verification score is the mean divergence over aligned unresolved positions:
\begin{equation}
    D_{\mathrm{JS}}^s
    =
    \frac{1}{|\mathcal{A}_s|}
    \sum_{j\in\mathcal{A}_s} d_j.
\end{equation}
The expanded branch is accepted only if
\begin{equation}
    D_{\mathrm{JS}}^s < \tau_{\mathrm{JS}},
\end{equation}
where $\tau_{\mathrm{JS}}$ is a fixed threshold. A low score indicates that insertion changes the predictive distributions at aligned unresolved positions only slightly on average. When the score exceeds the threshold, the proposed expansion is rejected. Because the score is averaged across positions, acceptance does not guarantee that every individual prediction is preserved.

\begin{algorithm}[t]
\caption{\textsc{\methodname{} Decoding with Verified Expansion}}
\label{alg:carve}
\small
\begin{algorithmic}[1]
\Require prompt $x_{\mathrm{p}}$, initial length $L_0$, maximum length $L_{\max}$, steps $T$, reverse denoising schedule $\{\tau_s\}_{s=0}^{T}$, insertion size $k$, threshold $\tau_{\mathrm{JS}}$, interval $I$
\State $x^0\gets \masktoken^{L_0}$
\For{$s=0,\ldots,T-1$}
    \If{$x^s$ contains no mask tokens}
        \State \Return decoded response from $x^s$
    \EndIf

    \State $L_s\gets |x^s|$
    \State $x_{\mathrm{B}}^s\gets x^s$
    \State $F_{\mathrm{B}}^s\gets
        f_\theta(x_{\mathrm{p}},x_{\mathrm{B}}^s,\tau_s)$
    \State $(x_{\mathrm{C}}^s,F_{\mathrm{C}}^s)
        \gets
        (x_{\mathrm{B}}^s,F_{\mathrm{B}}^s)$

    \If{$s \bmod I=0$ \text{and} $L_s<L_{\max}$}
        \State choose uncertainty anchor $c_s$
        \State $k_s\gets \min(k,L_{\max}-L_s)$
        \State $x_{\mathrm{E}}^s
            \gets
            \operatorname{Insert}(x_{\mathrm{B}}^s,c_s,k_s)$
        \State $F_{\mathrm{E}}^s
            \gets
            f_\theta(x_{\mathrm{p}},x_{\mathrm{E}}^s,\tau_s)$
        \State compute $D_{\mathrm{JS}}^s$ on aligned unresolved positions

        \If{$D_{\mathrm{JS}}^s<\tau_{\mathrm{JS}}$}
            \State $(x_{\mathrm{C}}^s,F_{\mathrm{C}}^s)
                \gets
                (x_{\mathrm{E}}^s,F_{\mathrm{E}}^s)$
        \EndIf
    \EndIf

    \State reveal selected masks in $x_{\mathrm{C}}^s$ using $F_{\mathrm{C}}^s$
    \State set $x^{s+1}$ to the result

    \If{EOS is committed at position $e_s$ before the canvas end}
        \State $\bar{x}^{s+1}\gets x^{s+1}_{1:e_s}$
        \State $\bar{F}^{s+1}\gets f_\theta(x_{\mathrm{p}},\bar{x}^{s+1},\tau_{s+1})$
        \State fill all remaining masks in $\bar{x}^{s+1}$ by greedily using argmax $\bar{F}^{s+1}$
        \State \Return decoded response from $\bar{x}^{s+1}$
    \EndIf
\EndFor
\State \Return decoded response from final canvas
\end{algorithmic}
\end{algorithm}

\paragraph{Branch selection.}
After verification, decoding continues on exactly one branch. We define the \emph{chosen branch} as
\begin{equation}
(x_{\mathrm{C}}^s,F_{\mathrm{C}}^s)
=
\begin{cases}
(x_{\mathrm{E}}^s,F_{\mathrm{E}}^s),
& D_{\mathrm{JS}}^s<\tau_{\mathrm{JS}},\\
(x_{\mathrm{B}}^s,F_{\mathrm{B}}^s),
& \text{otherwise}.
\end{cases}
\label{eq:branch_selection}
\end{equation}
This branch-consistent design is important: \methodname{} never commits tokens using logits from a canvas different from the one being updated.

\paragraph{Expansion interval.}
The expansion is controlled by a hyperparameter $I$, called the expansion interval. Rather than branching out at every denoising step, \methodname{} attempts expansion once every $I$ steps, as long as the active canvas has not reached $L_{\max}$. A standard decoding step requires one model forward pass. A step with an expansion attempt requires one additional forward pass for the expanded branch. Thus, before the canvas reaches $L_{\max}$, the average number of forward passes per decoding step is approximately $1+1/I$ times that of the underlying decoder; the corresponding FLOP overhead depends on the canvas length.

\paragraph{EOS-triggered cropping and greedy completion.}
\methodname{} uses EOS crop as its default stopping rule. After each reveal step, we check whether an end-of-sequence token has been committed before the end of the active response canvas. When this occurs, all positions to the right of the first EOS are discarded, since they would not contribute to the decoded answer. If the remaining prefix still contains mask tokens, the model performs one final forward pass on the cropped canvas and fills every unresolved position greedily by argmax. The resulting prefix is returned as the final response. This makes stopping consistent with answer extraction while avoiding additional denoising steps on tokens that would be discarded.

\section{Experiments}
\label{sec:experiments}

\begin{table*}[t]
\centering
\small
\begin{tabular}{llccccc}
\toprule
\textbf{Model} & \textbf{Method} & \textbf{HumanEval} & \textbf{MBPP} & \textbf{MATH-500} & \textbf{GSM8K} & \textbf{Average} \\
\midrule

\multirow{3}{*}{Dream-v0-Instruct-7B}
& Baseline
& 55.49
& \textbf{60.60}
& 39.60
& \textbf{79.53}
& 58.80 \\

& DAEDAL
& \textbf{59.76} \baseDelta{+4.27}
& 56.40 \baseDelta{$-4.20$}
& \textbf{40.80} \baseDelta{+1.20}
& 76.04 \baseDelta{$-3.49$}
& 58.25 \baseDelta{$-0.55$} \\

& \methodname{}
& \textbf{59.76} \cardelta{+4.27}
& 60.40 \carneg{$-0.20$}
& 40.20 \cardelta{+0.60}
& 78.54 \carneg{$-0.99$}
& \textbf{59.73} \cardelta{+0.92} \\

\midrule

\multirow{3}{*}{LLaDA-1.5}
& Baseline
& 46.95
& 38.20
& 43.00
& 83.09
& 52.81 \\

& DAEDAL
& 46.34 \baseDelta{$-0.61$}
& \textbf{39.80} \baseDelta{+1.60}
& 42.00 \baseDelta{$-1.00$}
& \textbf{83.85} \baseDelta{+0.76}
& 53.00 \baseDelta{+0.19} \\

& \methodname{}
& \textbf{50.61} \cardelta{+3.66}
& 37.80 \carneg{$-0.40$}
& \textbf{43.40} \cardelta{+0.40}
& 83.55 \cardelta{+0.46}
& \textbf{53.84} \cardelta{+1.03} \\

\midrule

\multirow{3}{*}{LLaDA-8B-Instruct}
& Baseline
& 46.95
& 38.80
& 43.20
& 81.88
& 52.71 \\

& DAEDAL
& 46.34 \baseDelta{$-0.61$}
& \textbf{39.20} \baseDelta{+0.40}
& \textbf{47.00} \baseDelta{+3.80}
& 81.12 \baseDelta{$-0.76$}
& \textbf{53.42} \baseDelta{+0.71} \\

& \methodname{}
& \textbf{48.17} \cardelta{+1.22}
& \textbf{39.20} \cardelta{+0.40}
& 43.40 \cardelta{+0.20}
& \textbf{82.03} \cardelta{+0.15}
& 53.20 \cardelta{+0.49} \\

\bottomrule
\end{tabular}
\caption{
Main results on code and mathematical reasoning benchmarks.
Scores are reported as percentages. HumanEval and MBPP use pass@1; MATH-500 and GSM8K use exact-match accuracy after answer extraction. ``Average'' is the unweighted arithmetic mean of the four task scores. Deltas are percentage-point differences relative to the baseline for the same model. Averages and deltas are computed from unrounded scores.
Boldface marks the best result, including ties, within each model block.
}
\label{tab:main_results}
\end{table*}
We evaluate \methodname{} on code generation and mathematical reasoning benchmarks using three instruction-tuned masked diffusion language models. We also report DAEDAL \cite{daedal} as a training-free variable-length decoding baseline. In addition to task accuracy, we measure the inference cost of each method using forward-pass FLOPs. 

\subsection{Models}

We evaluate Dream-v0-Instruct-7B, LLaDA-1.5, and LLaDA-8B-Instruct. Dream uses full-canvas masked diffusion decoding, while the LLaDA models use the LLaDA-family blockwise decoding setup, with block size 32 for LLaDA-1.5 and block size 64 for LLaDA-8B-Instruct.

\begin{figure*}[t]
    \centering
    \includegraphics[width=\textwidth]{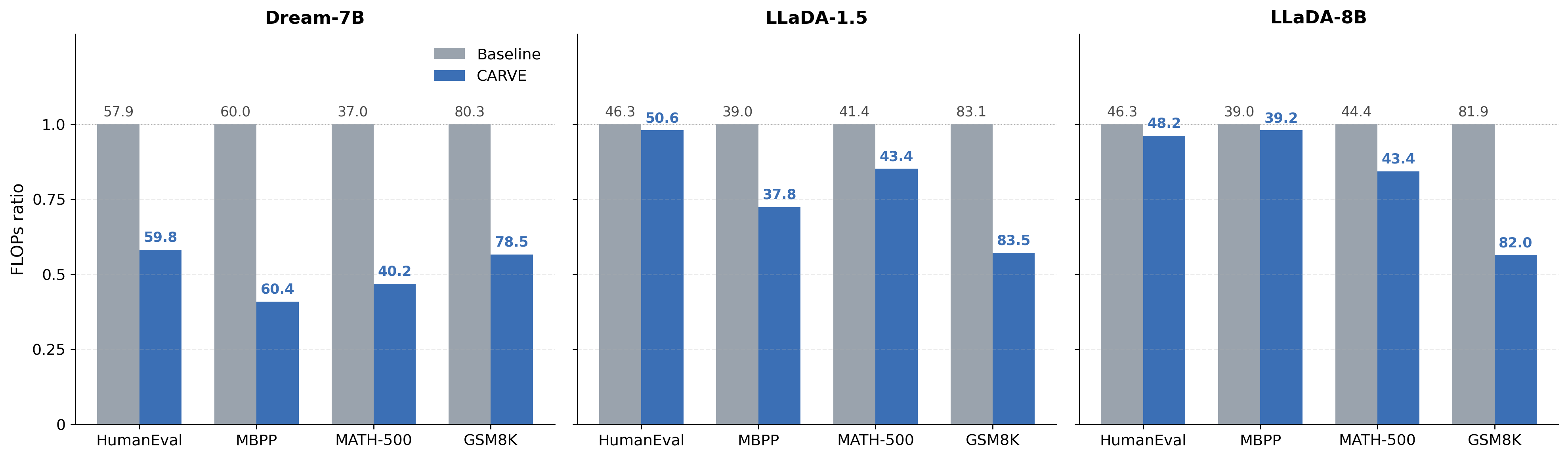}
    \caption{
    Per-task FLOPs ratio of \methodname{} relative to the fixed-length baseline.
    A value below $1.0$ means that \methodname{} uses fewer FLOPs than the baseline.
    Although \methodname{} performs an additional forward pass when it branches into an expanded canvas, EOS cropping and adaptive reveal often offset this cost, making the final decoding trajectory more efficient.
    }
    \label{fig:per_task_flops}
\end{figure*}

\subsection{Benchmarks}

We evaluate \methodname{} on four benchmarks: HumanEval \cite{humaneval}, MBPP \cite{mbpp}, MATH-500 \cite{math}, and GSM8K \cite{gsm8k}. Performance on HumanEval and MBPP is measured using pass@1, while performance on MATH-500 and GSM8K is measured using exact-match accuracy after answer extraction.

\subsection{Compared Methods}

We compare three decoding methods. \textbf{Baseline} denotes the standard fixed-length masked diffusion decoder for each model. \textbf{DAEDAL} is a training-free variable-length algorithm that expands the response canvas using internal confidence signals. Since DAEDAL is originally designed for LLaDA-style decoding, we also adapt it to Dream for a consistent comparison. \textbf{\methodname{}} is our training-free verified expansion algorithm. For all \methodname{} runs, we initialize the canvas at $L_0=L_{\max}/2$ and allow it to grow up to $L_{\max}$.

\subsection{Results}

Table~\ref{tab:main_results} shows that \methodname{} increases the unweighted four-benchmark average relative to fixed-length decoding by $0.92$, $1.03$, and $0.49$ percentage points on Dream-v0-Instruct-7B, LLaDA-1.5, and LLaDA-8B-Instruct, respectively. The per-task effects are mixed: \methodname{} improves 9 of the 12 model--benchmark pairs, with its largest gain on HumanEval for Dream ($+4.27$ points). DAEDAL attains the higher average on LLaDA-8B-Instruct ($53.42$ versus $53.20$).

This pattern of consistent improvements across diverse benchmarks for all evaluated model families suggests that verified expansion is both task-agnostic and model-agnostic. Rather than relying on a benchmark-specific length heuristic, \methodname{} uses the model's own predictive stability to decide when expansion is safe. The main exception is MATH-500 for LLaDA-8B-Instruct, where DAEDAL achieves the highest score; nevertheless, \methodname{} remains competitive on the aggregate and provides the most consistent gains across code and reasoning tasks.

Figure~\ref{fig:per_task_flops} shows that these accuracy gains do not come from simply spending more computation. Across models and tasks, \methodname{} often uses substantially fewer FLOPs than the fixed-length baseline. This may appear counterintuitive because the method branches out and performs an additional forward pass at any given expansion step. In practice, however, accepted expansions are controlled by the JS divergence, and EOS cropping removes suffix positions that would be discarded during answer extraction. As a result, the extra cost of branching is often balanced, and sometimes outweighed, by shorter effective decoding trajectories. Thus, \methodname{} improves accuracy while remaining computationally efficient.

\section{Ablations}
\subsection{Adaptive reveal rule}
We compare the adaptive reveal schedule used in \methodname{} against Dream's original commit schedule. Dream's original schedule reveals
\begin{equation}
    n_s
    =
    \left\lfloor
    m_s\left(1-\frac{\tau_{s+1}}{\tau_s}\right)
    \right\rfloor,
\end{equation}
where $m_s$ is the number of remaining masked positions and
$\{\tau_s\}_{s=0}^{T}$ is the reverse denoising schedule. This schedule can spend late denoising steps without revealing any token when the reveal count rounds to zero. We instead use
\begin{equation}
    n_s =
    \left\lceil
    \frac{m_s}{T-s}
    \right\rceil,
\end{equation}
which distributes the remaining masks across the remaining step budget and guarantees progress at each step. We use the same adaptive rule for the LLaDA backend. Holding the rest of the Dream configuration fixed, the adaptive schedule improves average accuracy from $58.17$ to $59.73$ while reducing the average number of forward passes from $200.9$ to $148.9$.

\begin{table}[t]
\centering
\small
\begin{tabular}{lcccc}
\toprule
\textbf{Schedule} & \textbf{HE} & \textbf{MBPP} & \textbf{MATH} & \textbf{GSM8K} \\
\midrule
Original
& 54.27
& \textbf{61.20}
& 38.80
& 78.39 \\
Adaptive
& \textbf{59.76}
& 60.40
& \textbf{40.20}
& \textbf{78.54} \\
$\Delta$
& \cardelta{+5.49}
& \carneg{$-0.80$}
& \cardelta{+1.40}
& \cardelta{+0.15} \\
\bottomrule
\end{tabular}
\caption{
Accuracy ablation of the adaptive reveal schedule on Dream-v0-Instruct-7B.
The $\Delta$ row reports percentage-point differences relative to the
original schedule.
}
\label{tab:adaptive_schedule_accuracy}
\end{table}

\begin{table}[t]
\centering
\small
\begin{tabular}{@{}lccccc@{}}
\toprule
\textbf{Schedule} & \textbf{HE} & \textbf{MBPP} & \textbf{MATH} &
\textbf{GSM8K} & \textbf{Avg.} \\
\midrule
Original
& 100.7
& 151.0
& 344.7
& 207.3
& 200.9 \\
Adaptive
& \textbf{74.5}
& \textbf{109.3}
& \textbf{261.8}
& \textbf{150.1}
& \textbf{148.9} \\
\bottomrule
\end{tabular}
\caption{
Forward-pass ablation of the adaptive reveal schedule on Dream-v0-Instruct-7B.
Lower is better; values are averaged per sample.
}
\label{tab:adaptive_schedule_fwds}
\end{table}

\begin{figure*}[t]
    \centering
    \includegraphics[width=\textwidth]{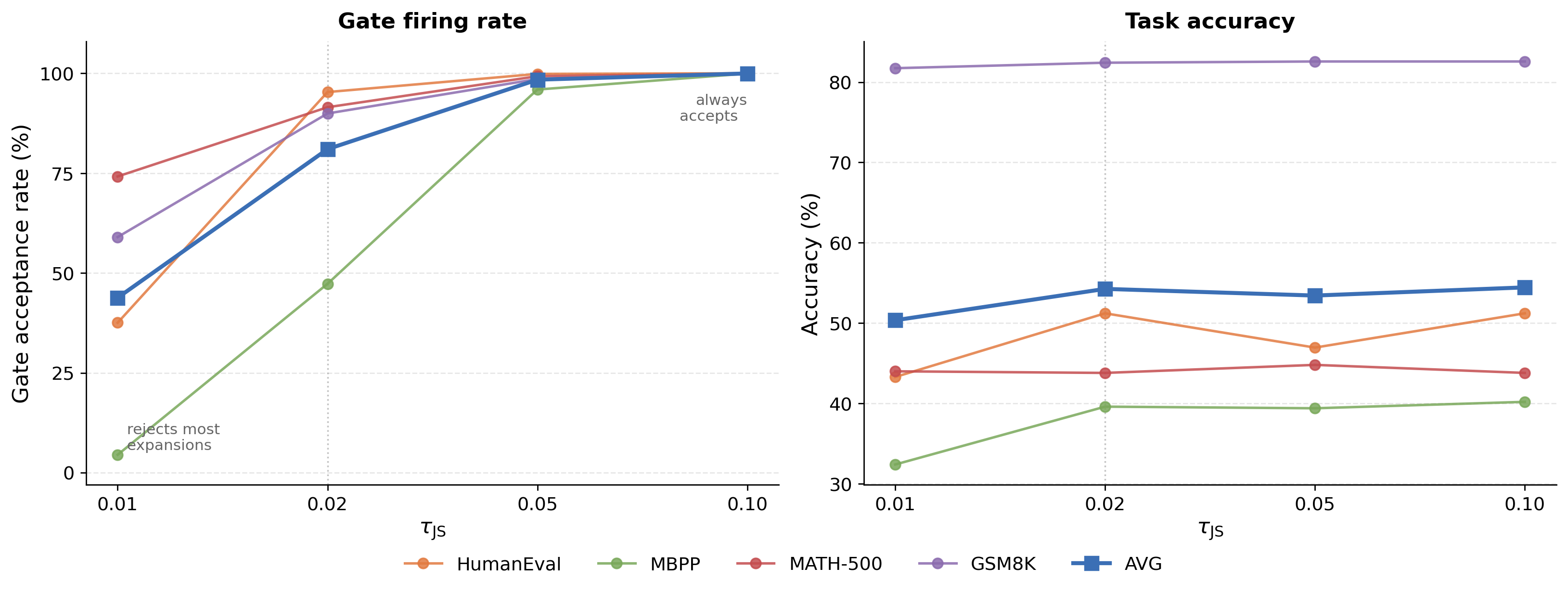}
    \caption{
    Effect of the JS acceptance threshold $\tau_{\mathrm{JS}}$ on LLaDA-8B.
    Left: percentage of accepted expansions.
    Right: task accuracy under the same thresholds.
    Very small thresholds reject too many expansions, while large thresholds accept almost all proposals.
    We use $\tau_{\mathrm{JS}}=0.02$, which reaches the accuracy plateau while keeping \methodname{} selective.
    }
    \label{fig:js_threshold_ablation}
\end{figure*}

\subsection{Effect of the JS threshold}
We ablate the expansion threshold $\tau_{\mathrm{JS}}$ on the LLaDA-8B setting by sweeping
$\tau_{\mathrm{JS}}\in\{0.01,0.02,0.05,0.10\}$. Figure~\ref{fig:js_threshold_ablation} reports both the fraction of accepted expansions and the resulting task accuracy. The threshold controls a clear trade-off. When $\tau_{\mathrm{JS}}=0.01$, CARVE is overly conservative: many expansions are rejected, especially on MBPP, which coincides with lower accuracy, consistent with the canvas not growing enough. Conversely, larger thresholds such as $0.05$ and $0.10$ accept nearly all proposals, making the verification step almost vacuous and moving the method toward an always-expand setting.

We use $\tau_{\mathrm{JS}}=0.02$ as the default because it is the smallest tested threshold at which average accuracy reaches its plateau while CARVE remains selective. At this value, \methodname{} preserves the average accuracy of more permissive thresholds but still rejects a nontrivial fraction of proposed insertions. This supports the role of the JS divergence as a meaningful stability check rather than a constant expansion rule.

\subsection{Insertion Mode}
We compare the uncertainty-guided insertion rule used by \methodname{} with a simpler tail-insertion strategy. Table~\ref{tab:insertion_mode_ablation} reports the average score of the best configuration found for each mode and model. Mid-insert performs best on Dream and LLaDA-1.5, suggesting that placing new masks near uncertain regions can help the model allocate capacity where the current canvas is under-specified. Tail insertion, however, remains competitive and is strongest on LLaDA-8B, indicating that the optimal insertion location can depend on the backbone and decoding dynamics. We therefore use mid-insert as the default canvas-aware rule, while treating tail insertion as a strong, simpler alternative.
\begin{table}[h]
\centering
\small
\setlength{\tabcolsep}{6pt}
\begin{tabular}{lcc}
\toprule
\textbf{Model} & \textbf{Mid-insert} & \textbf{Tail} \\
\midrule
Dream-7B & \textbf{59.73} & 59.00 \\
LLaDA-1.5 & \textbf{53.84} & 52.88 \\
LLaDA-8B & 53.20 & \textbf{54.29} \\
\bottomrule
\end{tabular}
\caption{
Insertion-mode ablation. We compare average task accuracy for mid-insert and tail insertion using the best configuration found for each mode and model.
}
\label{tab:insertion_mode_ablation}
\end{table}

\section{Discussion}
\label{sec:discussion}

\methodname{} expands the canvas only when the proposed insertion produces a small mean distribution shift at aligned unresolved positions. This stability criterion remains permissive enough to allow useful growth in the evaluated settings. Across models, \methodname{} consistently improves average accuracy over fixed-length baselines while often using less compute than fixed-length decoding.

The results suggest that effective length control for masked diffusion LMs is not only about adding more masks, but about adding them without destabilizing the denoising state. JS divergence provides this check by accepting an expanded branch only when the model's predictions remain stable after insertion. The ablations support this view: too small a threshold rejects useful growth, while too large a threshold makes \methodname{} nearly equivalent to always expanding.

Finally, the compute results show that the extra forward passes introduced by branching do not necessarily translate into higher total cost. In \methodname{}, branching is paired with EOS crop: once the model commits an end-of-sequence token, decoding stops on the useful prefix instead of continuing to refine suffix positions that would be discarded. Together with the adaptive reveal schedule, this offsets much of the cost of expansion attempts. As a result, \methodname{} improves the accuracy--compute trade-off using only inference-time changes, without retraining or modifying the underlying diffusion model.

\section{Conclusion}

We introduced \methodname{} (\textbf{C}ounterfactual-\textbf{A}ware \textbf{R}eveal with \textbf{V}erified \textbf{E}xpansion), a training-free variable-length decoding algorithm for masked diffusion language models. \methodname{} addresses the fixed-canvas limitation of dLLMs by branching into candidate expanded canvases and accepting an expansion only when it preserves the model's predictions on aligned unresolved positions. This turns length control into a counterfactual stability test rather than a pure confidence or EOS heuristic.

Across code-generation and mathematical-reasoning benchmarks, \methodname{} improves average performance over fixed-length baselines for each evaluated diffusion LM. Although branching adds forward passes, adaptive reveal and EOS cropping often offset this cost, yielding a better accuracy--compute trade-off than fixed-length decoding. Overall, our results show that pretrained masked diffusion LMs already contain useful signals for safe length adaptation, which can be exploited directly at inference time without retraining or architectural changes.

\section*{Limitations}
\label{sec:limitations}

\methodname{} currently inserts a fixed number of mask tokens at each accepted expansion. In our main configurations, this value is set to $k=16$. While this works well empirically, it does not adapt to the uncertainty or length requirements of each prompt. A natural direction for future work is to make insertion size adaptive, deciding not only where to expand the canvas but also how many new mask tokens should be added.

A second limitation is the alignment used by the JS divergence computation. \methodname{} compares predictions only on unresolved positions that already existed before insertion, while the newly inserted mask positions are excluded because they have no direct counterpart in the base canvas. This makes the verification step simple and well-defined, but it may be overly rigid. Future work could explore softer alignment or alternative divergence criteria that also account for the behavior of the newly inserted positions.

\bibliography{custom}

\appendix

\section{Experimental Settings}
\label{app:settings}

Table~\ref{tab:carve_configs} summarizes the task-specific decoding configurations for each model. For every model--benchmark pair, we initialize the canvas with \(L_0=L_{\max}/2\) masked positions and set the denoising budget to \(T=L_{\max}\). The table also reports the sampling temperature, block size, insertion size \(k\), uncertainty-window size \(W\), and expansion interval \(I\).

\begin{table}[t]
\centering
\small
\setlength{\tabcolsep}{3.2pt}
\resizebox{\columnwidth}{!}{
\begin{tabular}{lccccccc}
\toprule
\textbf{Model} &
\textbf{Canvas ($L_{\max}$)} &
\textbf{Steps} &
\textbf{Temp.} &
\textbf{Block} &
\textbf{$k$} &
\textbf{$W$} &
\textbf{$I$} \\
\midrule
Dream-7B
& 128 / 256 / 512 / 256
& $L_{\max}$
& 0.04
& full
& 16
& 12
& 16 \\
LLaDA-1.5
& 512 / 512 / 512 / 512
& $L_{\max}$
& 0.05
& 32
& 16
& 4
& 1 \\
LLaDA-8B
& 512 / 256 / 512 / 512
& $L_{\max}$
& 0.03
& 64
& 16
& 8
& 1 \\
\bottomrule
\end{tabular}
}
\caption{
Main hyperparameters used for the evaluations. The canvas \& $L_{\max}$ column reports HE / MBPP / MATH / GSM8K. The number of denoising steps is set equal to the task-specific maximum canvas length.}
\label{tab:carve_configs}
\end{table}
\section{Hardware}
\label{app:hardware}

All experiments used AMD MI210 GPUs and consumed approximately 42 aggregate GPU-days. Runs used at most eight GPUs concurrently.

\section{Isolating the Contribution of Each Component}
\label{app:component_ablation}

We decompose \methodname{} into its main components and evaluate controlled variants of the decoder. For each model and benchmark, all rows use the same prompts, maximum canvas length, reveal rule, and evaluation metric. The rows differ only in the decoding components enabled.

\paragraph{Configurations.}
\textit{Fixed baseline} denotes standard fixed-length decoding at $L_{\max}$.
\textit{+ adaptive reveal} uses the full canvas at $L_{\max}$ with the adaptive reveal rule, but without expansion or EOS cropping.
\textit{Full canvas + EOS-crop} adds EOS cropping to the previous setting.
\textit{Fixed-$L_0$ + EOS-crop} fixes the canvas at $L_0=L_{\max}/2$, disables expansion, and uses EOS cropping.
\textit{Always-expand} inserts new masks every $I$ steps unconditionally, with the verification forward removed; reveal decisions are therefore based on the pre-insertion forward.
\textit{\methodname{}} is the full method.
\begin{table*}[t]
\centering
\small
\setlength{\tabcolsep}{4.5pt}
\begin{tabular}{llcccccc}
\toprule
\textbf{Model} &
\textbf{Configuration} &
\textbf{HumanEval} &
\textbf{MBPP} &
\textbf{GSM8K} &
\textbf{MATH-500} &
\textbf{Avg.} &
\textbf{FLOPs$\times$} \\
\midrule

\multirow{6}{*}{Dream}
& Fixed baseline
& 55.49 & 60.60 & 79.53 & 39.60 & 58.80 & 1.00 \\
& + adaptive reveal
& 55.49 & 60.60 & 79.53 & 39.60 & 58.80 & 1.00 \\
& Full canvas + EOS-crop
& 55.49 & 60.60 & 79.53 & 39.60 & 58.80 & 0.49 \\
& Fixed-$L_0$ + EOS-crop
& 46.34 & 59.00 & 68.39 & 38.40 & 53.03 & 0.32 \\
& Always-expand
& 59.15 & 61.40 & 78.39 & 40.40 & 59.84 & 0.48 \\
& \textbf{\methodname{}}
& 59.76 & 60.40 & 78.54 & 40.20 & 59.73 & 0.51 \\

\midrule

\multirow{6}{*}{LLaDA-8B}
& Fixed baseline
& 46.95 & 38.80 & 81.88 & 43.20 & 52.71 & 1.00 \\
& + adaptive reveal
& 46.95 & 38.80 & 81.88 & 43.20 & 52.71 & 1.00 \\
& Full canvas + EOS-crop
& 46.95 & 38.80 & 81.88 & 43.20 & 52.71 & 0.83 \\
& Fixed-$L_0$ + EOS-crop
& 35.98 & 30.20 & 81.58 & 39.60 & 46.84 & 0.31 \\
& Always-expand
& 45.73 & 39.00 & 82.56 & 42.20 & 52.37 & 0.82 \\
& \textbf{\methodname{}}
& 48.17 & 39.20 & 82.03 & 43.40 & 53.20 & 0.85 \\

\midrule

\multirow{6}{*}{LLaDA-1.5}
& Fixed baseline
& 46.95 & 38.20 & 83.09 & 43.00 & 52.81 & 1.00 \\
& + adaptive reveal
& 46.95 & 38.20 & 83.09 & 43.00 & 52.81 & 1.00 \\
& Full canvas + EOS-crop
& 46.95 & 38.20 & 83.09 & 43.20 & 52.86 & 0.77 \\
& Fixed-$L_0$ + EOS-crop
& 38.41 & 39.40 & 81.80 & 39.00 & 49.65 & 0.30 \\
& Always-expand
& 48.17 & 37.60 & 83.93 & 43.80 & 53.38 & 0.75 \\
& \textbf{\methodname{}}
& 50.61 & 37.80 & 83.55 & 43.40 & 53.84 & 0.78 \\

\bottomrule
\end{tabular}
\caption{
Component-wise ablation across the three evaluated backends. FLOPs are normalized by the fixed-length baseline for the same model and benchmark. The ``+ adaptive reveal'', ``Full canvas + EOS-crop'', and ``Fixed-$L_0$ + EOS-crop'' rows use no expansion. The ``Always-expand'' row removes the verification forward pass and inserts unconditionally.
}
\label{tab:component_ladder_all}
\end{table*}

\paragraph{Discussion.}
These controlled ablations show that the gains of \methodname{} do not come from a single independent shortcut. Adding adaptive reveal on top of vanilla decoding leaves performance unchanged, as expected: with a fixed canvas, both use the same position-ranking criterion but different reveal-count schedules. Its main role is to make a growing canvas usable: once new masks are inserted, the decoder needs a reveal schedule that can keep pace with the changing number of unresolved positions.

EOS cropping behaves differently. It is primarily an efficiency mechanism: on a full canvas, it approximately preserves accuracy while reducing computation spent on suffix positions that would be discarded after EOS. However, cropping alone is not enough. When the canvas is fixed at $L_0=L_{\max}/2$ and cannot grow, the decoder becomes much cheaper but loses substantial accuracy, showing that a small initial canvas must be paired with a mechanism for allocating additional space.

The always-expand variant further clarifies the role of verification. On the LLaDA backends, removing the verification forward pass reduces accuracy relative to \methodname{} while saving only a small amount of compute. On Dream, the two variants are close. This suggests that unconditional growth can sometimes be sufficient, but is not a reliable replacement for verified expansion across backends. Moreover, this control removes not only the accept/reject decision, but also the forward pass from which an accepted expanded branch reveals its tokens. Without that pass, tokens are revealed from predictions computed before the canvas was enlarged.

\section{Adaptive Reveal Under a Growing Canvas}
\label{app:adaptive_growing}

Appendix~\ref{app:component_ablation} shows that the adaptive reveal rule is not a major source of fixed-canvas accuracy gains on a fixed canvas. Its role in \methodname{} is operational: when the canvas grows, newly inserted masks must also be revealed within the remaining denoising budget.

Dream's original schedule reveals
\[
n_s =
\left\lfloor
M_s
\left(
1-\frac{\tau_{s+1}}{\tau_s}
\right)
\right\rfloor,
\]
where $M_s$ is the number of remaining masked positions and $\{t_s\}_{s=0}^{T}$ is the denoising time grid. This value can round to zero, causing a model forward pass to reveal no tokens. \methodname{} instead uses
\[
n_s =
\left\lceil
\frac{M_s}{T-s}
\right\rceil,
\]
which distributes the remaining masked positions across the remaining denoising steps and guarantees at least one reveal per step while masks remain.

\begin{figure}[t]
    \centering
    \includegraphics[width=\columnwidth]{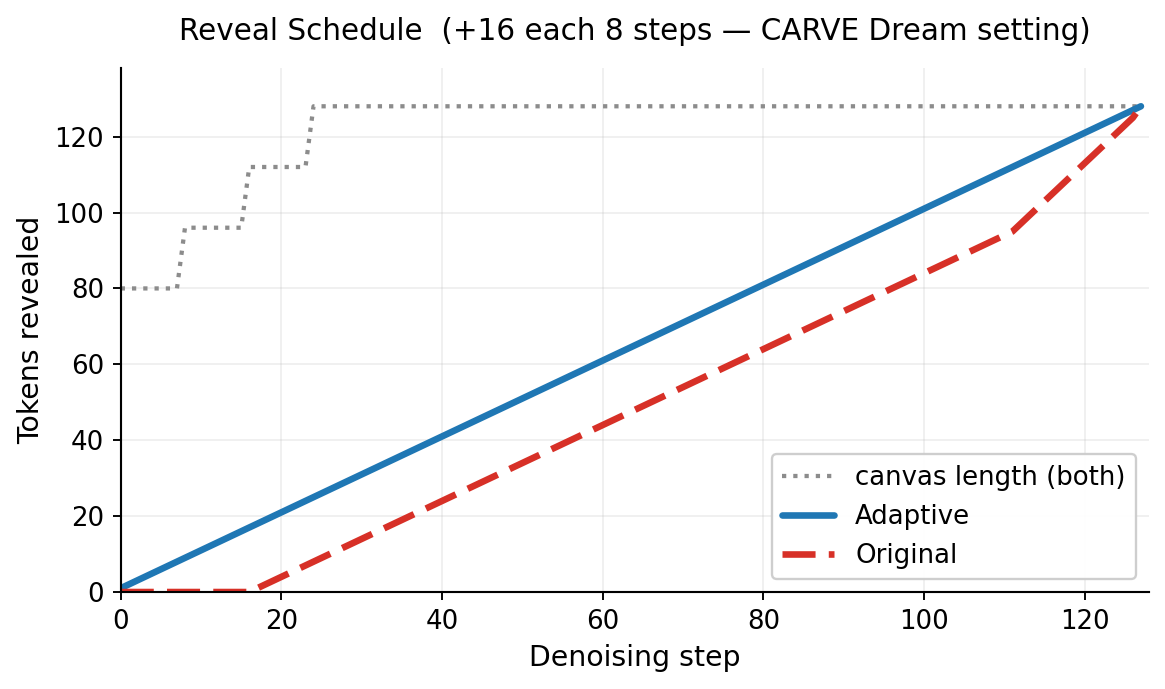}
    \caption{
    Reveal trace under a growing canvas in the Dream setting. The canvas grows by inserting 16 masks every 8 denoising steps until reaching $L_{\max}=128$. The original schedule, designed for a fixed canvas, initially reveals no tokens, then reveals many tokens late in the trajectory. The adaptive rule keeps reveal progress synchronized with the growing canvas.
    }
    \label{fig:reveal_trace_injection}
\end{figure}

Figure~\ref{fig:reveal_trace_injection} illustrates the mismatch between a fixed-canvas reveal schedule and a growing canvas. Under the original rule, the decoder spends early steps expanding the canvas without committing tokens, then has to reveal the remaining masks late in the trajectory. The adaptive rule avoids this stall by keeping the number of remaining masks aligned with the remaining step budget.


This same effect is reflected in Table~\ref{tab:adaptive_schedule_fwds}: the adaptive rule uses substantially fewer forward passes on Dream because it prevents reveal stalls under a growing canvas, reducing the average number of forward passes per sample from $200.9$ to $148.9$. Given this efficiency gain, we use the adaptive reveal rule by default for the LLaDA backends as well.

\section{Cost When the Response Fills the Canvas}
\label{app:longest_cost}

The average FLOPs reduction of \methodname{} partly comes from EOS cropping, which shortens the effective decoding trajectory. The least favorable case is therefore a response that fills the canvas, where cropping provides little or no benefit and the expansion overhead remains. To isolate this setting, we select the 5\% longest responses for each benchmark, i.e., examples that fill 99--100\% of the canvas, and compare \methodname{} against the fixed-length baseline on the same examples.

\begin{table}[H]
\centering
\small
\setlength{\tabcolsep}{4pt}
\resizebox{\columnwidth}{!}{
\begin{tabular}{lccc}
\toprule
\textbf{Benchmark} & \textbf{Dream} & \textbf{LLaDA-8B} & \textbf{LLaDA-1.5} \\
\midrule
HumanEval
& $1.01\times$ (99\%) & $1.02\times$ (100\%) & $1.02\times$ (100\%) \\
MBPP
& $1.01\times$ (100\%) & $1.04\times$ (100\%) & $1.02\times$ (100\%) \\
GSM8K
& $1.01\times$ (100\%) & $1.02\times$ (100\%) & $1.01\times$ (99\%) \\
MATH-500
& $0.98\times$ (100\%) & $1.02\times$ (100\%) & $1.02\times$ (100\%) \\
\bottomrule
\end{tabular}
}
\caption{
FLOPs ratio of \methodname{} relative to the fixed-length baseline on the 5\% longest responses per benchmark. Parentheses report the percentage of the canvas filled by the decoded response.
}
\label{tab:longest_flops}
\end{table}

As shown in Table~\ref{tab:longest_flops}, even in this adverse regime the overhead remains small. The largest observed cost is $1.04\times$ the fixed baseline, and most settings are within two percent of the baseline. This overhead is bounded by construction: expansion is attempted at most once every $I$ decoding steps and can therefore add at most $\lceil T/I\rceil$ expanded-branch forward passes over a $T$-step trajectory. Rejected proposals still incur this additional forward pass. EOS-triggered completion can add at most one further forward pass.

\section{Wall-Clock Throughput and Peak Memory}
\label{app:wallclock_memory}

FLOPs provide a hardware-independent proxy for inference cost. We additionally report wall-clock throughput and peak device memory on 8$\times$AMD MI210 GPUs. Both quantities are measured as ratios of \methodname{} to the fixed-length baseline.

\begin{table}[h]
\centering
\small
\begin{tabular}{lccc}
\toprule
\textbf{Benchmark} & \textbf{Dream} & \textbf{LLaDA-8B} & \textbf{LLaDA-1.5} \\
\midrule
HumanEval & 1.52 & 0.93 & 0.97 \\
MBPP      & 3.10 & 0.97 & 1.32 \\
GSM8K     & 1.75 & 1.63 & 1.60 \\
MATH-500  & 2.29 & 1.12 & 1.09 \\
\bottomrule
\end{tabular}
\caption{
Throughput ratio of \methodname{} relative to the fixed-length baseline, measured in tokens per second. Higher is better.
}
\label{tab:throughput_ratio}
\end{table}

\begin{table}[h]
\centering
\small
\begin{tabular}{lccc}
\toprule
\textbf{Benchmark} & \textbf{Dream} & \textbf{LLaDA-8B} & \textbf{LLaDA-1.5} \\
\midrule
HumanEval & 0.99 & 1.01 & 1.01 \\
MBPP      & 1.00 & 1.01 & 1.01 \\
GSM8K     & 1.01 & 1.01 & 1.01 \\
MATH-500  & 0.96 & 1.01 & 1.01 \\
\bottomrule
\end{tabular}
\caption{
Peak memory remains essentially unchanged empirically, with ratios
between $0.96$ and $1.01$ in the evaluated settings.
}
\label{tab:memory_ratio}
\end{table}

The throughput results broadly follow the FLOPs trends. \methodname{} is consistently faster on Dream and improves throughput on most LLaDA settings, with a few cases slightly below the fixed baseline when responses are long and cropping cannot offset the expansion forward passes. Peak memory remains essentially unchanged: \methodname{} never exceeds the same maximum canvas length used by the fixed-length decoder, and differs only in how that canvas is reached during decoding.

\section{Qualitative Examples}
\label{app:qualitative}

Figures~\ref{fig:qualitative_he14}--\ref{fig:qualitative_gsm0242} show qualitative examples from HumanEval and GSM8K using Dream-7B.

%
%
\providecommand{\qlhdr}{}
\definecolor{qlhdr}{HTML}{E6EAEE}   
\definecolor{qlrow}{HTML}{F4F5F6}   
\definecolor{qlmut}{HTML}{6B7280}   
\definecolor{qlok}{HTML}{1B7F4B}    
\definecolor{qlbad}{HTML}{C0392B}   
\providecommand{\qlpass}{\textcolor{qlok}{$\checkmark$}}
\providecommand{\qlfail}{\textcolor{qlbad}{$\times$}}
\providecommand{\qlstat}[1]{{\scriptsize\textcolor{qlmut}{#1}}}

\begin{figure*}[t]
\centering
{\footnotesize
\setlength{\tabcolsep}{5pt}
\renewcommand{\arraystretch}{1.25}
\begin{tabularx}{\textwidth}{@{}>{\raggedright\arraybackslash}p{2.15cm}>{\raggedright\arraybackslash}X@{}}
\toprule
\rowcolor{qlhdr}\textbf{Prompt} & \ttfamily from typing import List \newline
      \mbox{} \newline
      \mbox{} \newline
      def all\_prefixes(string: str) -\textgreater{} List[str]: \newline
      \mbox{}~~~~\textquotedbl{}\textquotedbl{}\textquotedbl{} Return list of all prefixes from shortest to longest of the input string \newline
      \mbox{}~~~~\textgreater{}\textgreater{}\textgreater{} all\_prefixes(\textquotesingle{}abc\textquotesingle{}) \newline
      \mbox{}~~~~[\textquotesingle{}a\textquotesingle{}, \textquotesingle{}ab\textquotesingle{}, \textquotesingle{}abc\textquotesingle{}] \newline
      \mbox{}~~~~\textquotedbl{}\textquotedbl{}\textquotedbl{} \\
\midrule
\rowcolor{qlrow}\textbf{Vanilla}~\qlfail \newline \qlstat{128 fwds} \newline \qlstat{11.95\,s} \newline \qlstat{len = 128} & \ttfamily \mbox{}~~~~prefixes = [] \newline
      \mbox{}~~~~for i in range(len(string) + 1): \newline
      \mbox{}~~~~~~~~prefixes.append(string[:i]) \newline
      \mbox{}~~~~return prefixes \\
\addlinespace[1pt]
\rowcolor{qlrow}\textbf{DAEDAL}~\qlpass \newline \qlstat{46 fwds} \newline \qlstat{4.91\,s} \newline \qlstat{len = 256} & \ttfamily prefixes = [] \newline
      \mbox{} \newline
      \mbox{}~~~~for i in range(len(string)): \newline
      \mbox{}~~~~~~~~prefixes.append(string[:i+1]) \newline
      \mbox{} \newline
      \mbox{}~~~~return prefixes \\
\addlinespace[1pt]
\rowcolor{qlrow}\textbf{CARVE}~\qlpass \newline \qlstat{30 fwds} \newline \qlstat{3.44\,s} \newline \qlstat{len = 96} & \ttfamily prefixes = [] \newline
      \mbox{}~~~~for i in range(len(string)): \newline
      \mbox{}~~~~~~~~prefixes.append(string[:i + 1]) \newline
      \mbox{}~~~~return prefixes \\
\bottomrule
\end{tabularx}}
\caption{Example HumanEval sample for Dream-7B methods (\texttt{HumanEval\_14}). \emph{fwds} is the number of model forward passes, \emph{len} the peak canvas length reached during decoding.}
\label{fig:qualitative_he14}
\end{figure*}

\begin{figure*}[t]
\centering
{\footnotesize
\setlength{\tabcolsep}{5pt}
\renewcommand{\arraystretch}{1.25}
\begin{tabularx}{\textwidth}{@{}>{\raggedright\arraybackslash}p{2.15cm}>{\raggedright\arraybackslash}X@{}}
\toprule
\rowcolor{qlhdr}\textbf{Prompt} & \ttfamily \mbox{} \newline
      def anti\_shuffle(s): \newline
      \mbox{}~~~~\textquotedbl{}\textquotedbl{}\textquotedbl{} \newline
      \mbox{}~~~~Write a function that takes a string and returns an ordered version of it. \newline
      \mbox{}~~~~Ordered version of string, is a string where all words (separated by space) \newline
      \mbox{}~~~~are replaced by a new word where all the characters arranged in \newline
      \mbox{}~~~~ascending order based on ascii value. \newline
      \mbox{}~~~~Note: You should keep the order of words and blank spaces in the sentence. \newline
      \mbox{} \newline
      \mbox{}~~~~For example: \newline
      \mbox{}~~~~anti\_shuffle(\textquotesingle{}Hi\textquotesingle{}) returns \textquotesingle{}Hi\textquotesingle{} \newline
      \mbox{}~~~~anti\_shuffle(\textquotesingle{}hello\textquotesingle{}) returns \textquotesingle{}ehllo\textquotesingle{} \newline
      \mbox{}~~~~anti\_shuffle(\textquotesingle{}Hello World!!!\textquotesingle{}) returns \textquotesingle{}Hello !!!Wdlor\textquotesingle{} \newline
      \mbox{}~~~~\textquotedbl{}\textquotedbl{}\textquotedbl{} \\
\midrule
\rowcolor{qlrow}\textbf{Vanilla}~\qlfail \newline \qlstat{128 fwds} \newline \qlstat{11.51\,s} \newline \qlstat{len = 128} & \ttfamily \mbox{}~~~~words = s.split() \newline
      \mbox{}~~~~ordered\_words = [] \newline
      \mbox{}~~~~for word in words: \newline
      \mbox{}~~~~~~~~ordered\_word = \textquotesingle{}\textquotesingle{}.join(sorted(word)) \newline
      \mbox{}~~~~~~~~ordered\_words.append(ordered\_word) \newline
      \mbox{}~~~~return \textquotesingle{} \textquotesingle{}.join(ordered\_words) \newline
      `{}`{}`{} \newline
      \mbox{} \newline
      You can test this function with the provided examples: \newline
      `{}`{}`{}python \newline
      assert anti\_shuffle(\textquotesingle{}Hi\textquotesingle{}) == \textquotesingle{}Hi\textquotesingle{} \newline
      assert anti\_shuffle(\textquotesingle{}hello\textquotesingle{}) == \textquotesingle{}ehllo\textquotesingle{} \newline
      assert anti\_shuffle(\textquotesingle{}Hello World!!!\textquotesingle{}) == \textquotesingle{}Hello !!!Wdlor\textquotesingle{} \\
\addlinespace[1pt]
\rowcolor{qlrow}\textbf{DAEDAL}~\qlpass \newline \qlstat{137 fwds} \newline \qlstat{15.27\,s} \newline \qlstat{len = 280} & \ttfamily words = s.split() \newline
      \mbox{}~~~~sorted\_words = [] \newline
      \mbox{} \newline
      \mbox{}~~~~for word in words: \newline
      \mbox{}~~~~~~~~sorted\_word = \textquotesingle{}\textquotesingle{}.join(sorted(word)) \newline
      \mbox{}~~~~~~~~sorted\_words.append(sorted\_word) \newline
      \mbox{} \newline
      \mbox{}~~~~return \textquotesingle{} \textquotesingle{}.join(sorted\_words) \newline
      `{}`{}`{} \newline
      \mbox{} \newline
      The function `{}anti\_shuffle`{} takes a string `{}s`{} as input and returns an ordered version of it. \newline
      \mbox{}~~~~It splits the string into words using the `{}split`{} method, and then iterates over each word \newline
      \mbox{}~~~~to sort its characters in ascending order using the `{}sorted`{} function. The sorted words are \newline
      \mbox{}~~~~then joined back together with spaces to form the final string. \\
\addlinespace[1pt]
\rowcolor{qlrow}\textbf{CARVE}~\qlpass \newline \qlstat{48 fwds} \newline \qlstat{4.40\,s} \newline \qlstat{len = 112} & \ttfamily words = s.split() \newline
      \mbox{}~~~~ordered\_words = [] \newline
      \mbox{}~~~~for word in words: \newline
      \mbox{}~~~~~~~~ordered\_word = \textquotesingle{}\textquotesingle{}.join(sorted(word)) \newline
      \mbox{}~~~~~~~~ordered\_words.append(ordered\_word) \newline
      \mbox{}~~~~return \textquotesingle{} \textquotesingle{}.join(ordered\_words) \\
\bottomrule
\end{tabularx}}
\caption{Example HumanEval sample for Dream-7B methods (\texttt{HumanEval\_86}). \emph{fwds} is the number of model forward passes, \emph{len} the peak canvas length reached during decoding.}
\label{fig:qualitative_he86}
\end{figure*}

\begin{figure*}[t]
\centering
{\footnotesize
\setlength{\tabcolsep}{5pt}
\renewcommand{\arraystretch}{1.25}
\begin{tabularx}{\textwidth}{@{}>{\raggedright\arraybackslash}p{2.15cm}>{\raggedright\arraybackslash}X@{}}
\toprule
\rowcolor{qlhdr}\textbf{Question} & \itshape John decides to do several activities while out on vacation.~~He spends 6 hours boating and half that time \newline
      \mbox{}~~~~swimming.~~He also watched 3 different shows which were 2 hours each.~~This was 30\% of the time he spent. \newline
      \mbox{}~~~~He spent 40\% of his time sightseeing.~~How much time did he spend sightseeing? \\
\rowcolor{qlhdr}\textbf{Answer} & \ttfamily 20 \\
\midrule
\rowcolor{qlrow}\textbf{Vanilla}~\qlfail \newline \qlstat{256 fwds} \newline \qlstat{69.01\,s} \newline \qlstat{len = 256} & \ttfamily He spent 6 hours boating so he spent 6 / 2 = \textless{}\textless{}6/2=3\textgreater{}\textgreater{}3 hours swimming. So he spent 6 + 3 = \newline
      \mbox{}~~~~\textless{}\textless{}6+3=9\textgreater{}\textgreater{}9 hours boating and swimming. So he spent 9 + 6 = \textless{}\textless{}9+6=15\textgreater{}\textgreater{}15 hours on those \newline
      \mbox{}~~~~activities. So he spent 15 / 30 = \textless{}\textless{}15/30=0.5\textgreater{}\textgreater{}0.5 of his time on those activities. So he \newline
      \mbox{}~~~~spent 1 / 0.5 = \textless{}\textless{}1/0.5=2\textgreater{}\textgreater{}2 hours on sightseeing. So he spent 2 / 40 = \textless{}\textless{}2/40=0.05\textgreater{}\textgreater{}0.05 \newline
      \mbox{}~~~~of his time sightseeing. So he spent 1 / 0.05 = \textless{}\textless{}1/0.05=20\textgreater{}\textgreater{}20 hours on sightseeing. So he \newline
      \mbox{}~~~~spent 20 * 40 = \textless{}\textless{}20*40=800\textgreater{}\textgreater{}800 minutes sightseeing. So he spent 800 / 60 = \newline
      \mbox{}~~~~\textless{}\textless{}800/60=13\textgreater{}\textgreater{}13 hours sightseeing. \newline
      \#\#\#\# 13 \\
\addlinespace[1pt]
\rowcolor{qlrow}\textbf{DAEDAL}~\qlfail \newline \qlstat{158 fwds} \newline \qlstat{46.00\,s} \newline \qlstat{len = 296} & \ttfamily He spent 6 hours boating so he spent 6 / 2 = \textless{}\textless{}6/2=3\textgreater{}\textgreater{}3 hours swimming. So he spent 6 + 3 = \newline
      \mbox{}~~~~\textless{}\textless{}6+3=9\textgreater{}\textgreater{}9 hours boating and swimming. So he spent 3 * 2 = \textless{}\textless{}3*2=6\textgreater{}\textgreater{}6 hours watching shows. \newline
      \mbox{}~~~~So he spent 9 + 6 = \textless{}\textless{}9+6=15\textgreater{}\textgreater{}15 hours on those activities. So he spent 100 - 30 = \newline
      \mbox{}~~~~\textless{}\textless{}100-30=70\textgreater{}\textgreater{}70\% of his time on those activities. So he spent 15 / .7 = \newline
      \mbox{}~~~~\textless{}\textless{}15/.7=21.428571428571428\textgreater{}\textgreater{}21.428571428571428 hours on his vacation. So he spent \newline
      \mbox{}~~~~21.428571428571428 * .4 = \textless{}\textless{}21.428571428571428*.4=8.571428571428571\textgreater{}\textgreater{}8.571428571428571 \newline
      \mbox{}~~~~hours sightseeing. \newline
      \#\#\#\# 8.57 \\
\addlinespace[1pt]
\rowcolor{qlrow}\textbf{CARVE}~\qlpass \newline \qlstat{146 fwds} \newline \qlstat{38.41\,s} \newline \qlstat{len = 256} & \ttfamily He spent 6 * 0.5 = \textless{}\textless{}6*0.5=3\textgreater{}\textgreater{}3 hours swimming. So he spent 6 + 3 = \textless{}\textless{}6+3=9\textgreater{}\textgreater{}9 hours boating \newline
      \mbox{}~~~~and swimming. So he spent 9 + 2 * 3 = \textless{}\textless{}9+2*3=15\textgreater{}\textgreater{}15 hours on those activities. So he spent \newline
      \mbox{}~~~~15 / 0.3 = \textless{}\textless{}15/0.3=50\textgreater{}\textgreater{}50 hours on vacation. So he spent 50 * 0.4 = \textless{}\textless{}50*0.4=20\textgreater{}\textgreater{}20 hours \newline
      \mbox{}~~~~sightseeing. \newline
      \#\#\#\# 20 \\
\bottomrule
\end{tabularx}}
\caption{Example GSM8K sample for Dream-7B methods (\texttt{GSM8K\_0353}). \emph{fwds} is the number of model forward passes, \emph{len} the peak canvas length reached during decoding.}
\label{fig:qualitative_gsm0353}
\end{figure*}

\begin{figure*}[t]
\centering
{\footnotesize
\setlength{\tabcolsep}{5pt}
\renewcommand{\arraystretch}{1.25}
\begin{tabularx}{\textwidth}{@{}>{\raggedright\arraybackslash}p{2.15cm}>{\raggedright\arraybackslash}X@{}}
\toprule
\rowcolor{qlhdr}\textbf{Question} & \itshape Mike was a pen pal with 5 people.~~He stopped being penpals with 2 of them.~~They each send 2 letters a week \newline
      \mbox{}~~~~that are 5 pages long.~~He responds in kind.~~He can write a page every 6 minutes.~~How many hours does \newline
      \mbox{}~~~~he spend writing a week? \\
\rowcolor{qlhdr}\textbf{Answer} & \ttfamily 3 \\
\midrule
\rowcolor{qlrow}\textbf{Vanilla}~\qlfail \newline \qlstat{256 fwds} \newline \qlstat{68.80\,s} \newline \qlstat{len = 256} & \ttfamily He stopped being penpals with 2 people so he is now penpals with 5 - 2 = \textless{}\textless{}5-2=3\textgreater{}\textgreater{}3 people. \newline
      They each send 2 letters a week that are 5 pages long so he gets 2 * 5 = \textless{}\textless{}2*5=10\textgreater{}\textgreater{}10 pages a \newline
      \mbox{}~~~~week. \newline
      He responds in kind so he writes 10 pages a week. \newline
      He can write a page every 6 minutes so he writes 10 * 6 = \textless{}\textless{}10*6=60\textgreater{}\textgreater{}60 minutes a week. \newline
      There are 60 minutes in an hour so he writes 60 / 60 = \textless{}\textless{}60/60=1\textgreater{}\textgreater{}1 hour a week. \newline
      \#\#\#\# 1 \\
\addlinespace[1pt]
\rowcolor{qlrow}\textbf{DAEDAL}~\qlfail \newline \qlstat{119 fwds} \newline \qlstat{34.59\,s} \newline \qlstat{len = 320} & \ttfamily He stopped being penpals with 2 people so he is now penpals with 5 - 2 = \textless{}\textless{}5-2=3\textgreater{}\textgreater{}3 people. \newline
      They each send 2 letters a week that are 5 pages long so that is 2 * 5 = \textless{}\textless{}2*5=10\textgreater{}\textgreater{}10 pages per \newline
      \mbox{}~~~~person. \newline
      He responds in kind so he writes 10 * 3 = \textless{}\textless{}10*3=30\textgreater{}\textgreater{}30 pages a week. \newline
      He can write a page every 6 minutes so he spends 30 * 6 = \textless{}\textless{}30*6=180\textgreater{}\textgreater{}180 minutes a week. \newline
      \#\#\#\# 180 \\
\addlinespace[1pt]
\rowcolor{qlrow}\textbf{CARVE}~\qlpass \newline \qlstat{129 fwds} \newline \qlstat{33.72\,s} \newline \qlstat{len = 256} & \ttfamily He stopped being penpals with 5 - 2 = \textless{}\textless{}5-2=3\textgreater{}\textgreater{}3 people. \newline
      He writes 3 * 2 = \textless{}\textless{}3*2=6\textgreater{}\textgreater{}6 letters a week. \newline
      He writes 6 * 5 = \textless{}\textless{}6*5=30\textgreater{}\textgreater{}30 pages a week. \newline
      He writes 30 * 6 = \textless{}\textless{}30*6=180\textgreater{}\textgreater{}180 minutes a week. \newline
      He writes 180 / 60 = \textless{}\textless{}180/60=3\textgreater{}\textgreater{}3 hours a week. \newline
      \#\#\#\# 3 \\
\bottomrule
\end{tabularx}}
\caption{Example GSM8K sample for Dream-7B methods (\texttt{GSM8K\_0242}). \emph{fwds} is the number of model forward passes, \emph{len} the peak canvas length reached during decoding.}
\label{fig:qualitative_gsm0242}
\end{figure*}




\end{document}